\def\year{2018}\relax
\documentclass[letterpaper]{article} 
\usepackage{aaai18}  
\usepackage{times}  
\usepackage{helvet}  
\usepackage{courier}  
\usepackage{url}  
\usepackage{graphicx}  

\usepackage{amsmath}
\usepackage[ruled,vlined]{algorithm2e}
\usepackage{color}
\usepackage{multirow}
 \usepackage{subcaption}
\newcommand{\prog}[1]{\mbox{{\small\sf #1}}} 
\newcommand{\initTS}{\prog{initialize}}
\newcommand{\updateTS}{\prog{update}}
\newcommand{\selectTS}{\prog{chooseArm}}

\begin{document}
%
\title{Learning Robust Search Strategies Using a Bandit-Based Approach}
\author{Wei Xia \and Roland H. C. Yap \\ School of Computing, National University of Singapore, Singapore \\ {\tt \{xiawei,ryap\}@comp.nus.edu.sg}
}
\maketitle

\begin{abstract}
Effective solving of constraint problems often requires choosing
good or specific search heuristics. 
However, choosing or designing a good search heuristic is
non-trivial and is often a manual process. 
In this paper, rather than manually choosing/designing search heuristics, 
we propose the use of bandit-based learning techniques to 
automatically select search heuristics. 
Our approach is online where the solver learns and 
selects from a set of heuristics during search. 
The goal is to obtain automatic search heuristics which give robust
performance.
Preliminary experiments show that our adaptive technique is more 
robust than the original search heuristics. 
It can also outperform the original heuristics.
\end{abstract}

\section{Introduction}
Constraint programming (CP) is successfully used in solving combinatorial problems.
In CP, problems are modeled as constraint satisfaction problems (CSP), often
NP hard.
Due to their intractability, CP solvers combine constraint solving
with a search strategy to instantiate variables.
A good search strategy can significantly reduce the size of the search space
giving faster problem solving.
Many variable search heuristics, e.g.~\textit{ddeg/dom}~\cite{bessiere96mac,Smith98ddeg}, 
\textit{wdeg/dom}~\cite{boussemart04boosting}, 
\textit{impact}~\cite{refalo04impact}, 
\textit{activity}~\cite{michel12activity}
and \textit{corr}~\cite{corr2017}, have been proposed.

However, choosing a heuristic from the
many existing variable ordering heuristics 
which works well for a particular problem instance 
or family is not simple. It may require expertise or experience.
A wrong (choice of) heuristic may mean that the solving time is slower
by several orders of magnitude, e.g.,
the \textit{activity} heuristic can be more than 100X faster than 
\textit{wdeg/dom} for some nurse-rostering problem instances, 
but can be slower for other benchmarks like radiation therapy
\cite{michel12activity}.
By the nature of heuristics, no heuristic always outperforms another.
In order to provide an efficient solution to a problem, 
specific search heuristics may need to be used, 
requiring considerable effort in choosing/designing
the heuristics for good performance.
A drawback of an individual heuristic is that it may only make a good decision at a certain solving state of a problem, e.g.~some heuristic may only perform well at a certain search depth, but not for the whole solving process. 

To address this challenge, we propose automatic and adaptive CSP 
search heuristics.
Our approach is motivated by the multi-armed bandit (MAB) 
problem~\cite{Gittins:1989} in reinforcement learning.
We consider the search heuristic choice as selecting dynamically from
several candidate variable ordering heuristics during search.
Each choice of search heuristic is akin to selecting an action
(arm) in the MAB. 
Failures from a search node are turned into rewards for the choice made
which affect subsequent choices.
The idea is that learning from the rewards of choosing a particular
heuristic will reduce making poor choices and in turn lead to a
search heuristic which is more robust.

We adapt two MAB algorithms, Thompson Sampling~\cite{mabts1933} and  UCB1 (upper confidence bound)~\cite{mabucb12002} with reward functions to select the ``best arm'', i.e.~``variable ordering heuristic''.
Our variable heuristics learn from information collected 
during the solving of the particular problem. 
Thus, learning is {\em online} and differs from
supervised learning approaches which require training examples and
an offline training solving phase. 
We study the performance of the original heuristics and our MAB-based heuristics
on a variety of benchmark families.
We also compare with a purely stochastic baseline method that randomly
selects a candidate heuristic at each node during search. 
Preliminary experiments show that our proposed adaptive learning techniques are more robust than the original heuristics with less variance for different classes of problems and instances.
Our online adaptive heuristics also outperforms the original heuristics 
on many problem instances.

Making CP solvers black box and robust is highly desirable. 
This paper is a step in this direction as the solver can determine 
the search strategy rather than being specified in the constraint
model while still giving good performance.

\subsection{Related work}\label{sec:ref}

Adaptive CSP search strategies using machine learning techniques have been studied in CP.
One distinguishing factor is whether an offline training phase is used.
Portfolio methods employ offline training, using the learned training
to select a solving strategy, which could be a search algorithm or a heuristic,
when solving a particular problem instance.
Some portfolio approaches are CPHydra~\cite{cphydra2008} and Proteus~\cite{proteus2014}. 
In addition,
some approaches learn and generate new solving strategies in the offline phase, e.g.~\cite{epstein2007,xu2009}.

Online learning approaches differ in that they do not
include a static learning phase before solving a problem.
For example, the Monte-Carlo tree search based method in \cite{mcstbandit2013} 
tries to expand the most promising nodes with online learning. 
The score function of the value heuristic is learned using a 
linear regression method in~\cite{Chu2015}. Bachiri et al.~\cite{Bachiri15} 
propose to learn the rewards of nodes and use the rewards to 
guide the search to backtrack to certain nodes. 
A recent work on adaptive search heuristics is the parallel strategies selection (PSS) approach~\cite{pss2016}.
PSS first decomposes the CSP into a large number sub-problems.
As the sub-problems are independent, parallelism can be readily used.
Sampling is used with parallel solving to select a heuristic and
the remaining sub-problems are solved in parallel with the heuristic.

The closest work is Balafrej et al.~\cite{paparrizou2015} which proposes
a MAB framework to select different levels of propagation during search.
They use the UCB1 algorithm to adaptively select the consistency level
at each node of the search tree.
In their experiments on binary CSPs, they show that learning can find when
higher consistencies than arc consistency should be employed during search.
Our work differs in that we adapt their MAB framework 
for the problem of selecting search heuristics dynamically 
and our experiments are on non-binary CSPs.
 
\section{Preliminaries}\label{sec:pre}

A constraint network $\mathcal{P}$ (CSP) is a pair $\mathcal{(X,C)}$ where
$\mathcal{X}$ is a set of $n$ variables $\{x_1, \ldots, x_n\}$ and
$\mathcal{C}$ a set of $e$ constraints $\{c_1,\ldots,c_e\}$. $D(x)$ is
the domain of $x \in \mathcal{X}$.  
Each $c \in \mathcal C$ involves two
components: a scope ($scp(c)$) which is an ordered subset of variables
of $\mathcal X$; and a relation over the scope ($rel(c)$).  Given
\emph{scp}($c$) = $\{x_{i_1}, \dots, x_{i_r}\}$, $rel(c)$ $\subseteq$
$\prod_{j=1}^{r}$ $D(x_{i_j})$ represents the set of satisfying
combinations of values for the variables in $scp(c)$.
We define {\em degree} of variable $x$ to be the number of constraints that $x$ belongs to. 
The \emph{arity} of $c$ is $|scp(c)|$.  
A binary CSP is of arity 2, while a non-binary CSP has constraints with arity $> 2$.
A \emph{solution} to
$\mathcal{P}$ is an assignment to all variables in $\mathcal{X}$ such that every constraint is satisfied.   

Constraint solvers typically explore the solution space 
by instantiating variables in some order. 
Usually, a variable ordering heuristic defines a score function, 
instantiating the variable with highest score at each search node.
Static variable ordering heuristics compute variable scores before search,
thus variable ordering is static. 
Dynamic ones update scores and tune the variable ordering dynamically during search. 
In practice, most of the successful variable ordering heuristics are dynamic ones, including \textit{ddeg/dom}~\cite{bessiere96mac,Smith98ddeg}, \textit{wdeg/dom}~\cite{boussemart04boosting}, \textit{impact}~\cite{refalo04impact}, and \textit{activity}~\cite{michel12activity}.
The \textit{ddeg/dom} and \textit{wdeg/dom} heuristics take the degrees and the current domain sizes of variables as parameters 
to the score functions. 
In \textit{ddeg/dom}, a variable's score is the value of its current degree divided by the variable's current domain size.
The current degree of a variable is the number of constraints, involving the variable, whose non-instantiated variables are more than one.
This is extended to weighted degree in
\textit{wdeg/dom}, a variable's score is the values of its weighted degree divided by variable's current domain size.
The weighted degree of a variable is the number of accumulated failures of the constraints which the variable belongs to.
The \textit{impact} heuristic selects the most influential variable which has made the most search space reduction in the space have been explored.
The \textit{activity} heuristic measures activity, i.e.~how often a variable's domain is filtered by constraint propagation, selecting the most active. 

The multi-armed bandit problem~\cite{Gittins:1989} comes from slot machines (one-armed bandit).
A player chooses a slot machine from multiple ones (multi-armed bandit) to maximize the total expected payoffs or rewards for a sequence of plays.
In MAB, an important consideration is the tradeoff between
{\em exploration} and {\em exploitation}.
An MAB algorithm should
exploit the actions with maximal rewards. However, without
exploring other actions enough, the algorithm may lose the
opportunity for finding better actions. Thus, an MAB algorithm
balances between exploration and exploitation.
One way is to minimize the cumulative regret.
Two of the successful and well-known MAB algorithms are the \textit{Thompson Sampling} (TS) algorithm~\cite{mabts1933} and the  \textit{Upper Confidence Bound algorithm} UCB1~\cite{mabucb12002}. 
Thompson sampling is one of the earliest algorithms and easy to implement. 
In practice, UCB1 is widely used for MAB due to its simplicity.
It guarantees a logarithmic increase in regret.  
We apply these algorithms to our problem
because of recent promising results~\cite{paparrizou2015,phillips2015efficient}
and due to their simplicity.
They can also be used as a standard baseline~\cite{mabucb12002,NIPS2011_4321}. 
 
\section{Multi-armed bandit for adaptive search}\label{sec:mab}
We consider the problem of selecting a variable ordering heuristic to pick 
which variable to explore in the search tree to be analogous to the
multi-armed bandit (MAB) problem. 
We map the automatic selection of variable ordering heuristics as 
a multi-armed bandit problem as follows.
We define a set of $\mathcal{K}$ arms $\{\mathcal{L}_1, \ldots,\mathcal{L}_k\}$ where each arm $\mathcal{L}_i$ corresponds to one candidate heuristic. 
MAB algorithms aim to maximize the total rewards and take actions based on 
the reward of each arm.
We can determine the rewards during search, thus,
a \textit{reward} $R_i(j)$ is associated with each arm $i\in[1,k]$ 
at each node $j$ in the search tree.
While solving CSPs, we would like to explore the solution space more
quickly.  One strategy is to try to make the search tree smaller.
Note that the propagators are fixed during search since the propagators
come from the constraints of the problem.
Thus, we define the rewards of candidate heuristics taken at each search node to be based on the number of children of the node. 

\subsection{MAB-based adaptive search framework}

In this paper, we propose a generic MAB-based search framework 
and adapt two specific MAB algorithms,
Thompson Sampling and Upper Confidence Bound 1,
to this framework.
Our MAB search framework adapts MAB algorithm
for the problem of dynamic selection of heuristics during backtracking
search in CSP solving as follows:
\begin{enumerate}
\item (\initTS) Initialize data structures of the MAB algorithm before search starts.\label{case:1}
\item (\selectTS) For each unexplored search node, first use the MAB algorithm to select a heuristic (an arm) and bind the selected heuristic to the node. 
Then use the selected heuristic to instantiate variables at the node. 
\item (\updateTS) Once search backtracks from a child node to its parent, which indicates that the child node is fully explored, update the mean rewards of the heuristic bound to the child node. 
\end{enumerate}

The main question is how to define the rewards for the MAB.
Our aim is to speed up solving by reducing the size of the 
explored search tree. 
As we select a heuristic at each search node, we will define the 
reward of the heuristic to be based on observations of the node. 
We propose to use the number of children of each search node to 
estimate the exploration of each search node since
search cost is usually correlated with the number of choices from a node.
Thus, we set the reward of the heuristic taken at a certain search node 
to be based on the number of children of that node. 

We emphasize that at each search node, only one heuristic is taken.
Thus, the MAB selection of the choice of heuristic is performed 
only once at a node.  
Some of the underlying heuristics used (in arm selection)
may the scores of variables in some accumulated fashion during search, e.g.~\textit{wdeg/dom} counts the accumulated number of failures.
So the scores of variables for all the relevant heuristics need to be maintained at each search node during search. 
When search backtracks from a child node $j$ to its parent node, 
we compute the rewards of the heuristic taken at node $j$ and update 
the rewards of the particular heuristic and other parameters depending 
of the particular MAB algorithm during the execution of the update step.  

\subsection{Thompson adaptive search}\label{sec:ts}
 
\begin{algorithm}[t]
\DontPrintSemicolon
\textbf{procedure} \initTS()\;  
\Begin{
\For{$i \in \{1 \ldots k \} $}{
$R_i^{ts} \leftarrow 0$, $R_{best}^{ts}[i] \leftarrow 0$ \;
$\alpha_i \leftarrow 1$, $\beta_i \leftarrow 1$\;
}
}

\textbf{procedure} \selectTS() \;
\Begin{
\For{$i \in \{1 \ldots k \} $}{
\tcp{sample from the distribution}
$\rho[i] \sim Beta(\alpha_i,\beta_i)$\; 
}
\textbf{return} arm $i$ s.t.~$\rho[i]=max\{\rho[1], \ldots, \rho[k]\}$\;
}

\textbf{procedure} \updateTS(i, r)\;
\Begin{
$R_i^{ts} \leftarrow r$ \;
\nl \eIf{$R_i^{ts} \geq R_{best}^{ts}[i]$}{ \label{alg:ts:l1}
$R_{best}^{ts}[i]\leftarrow R_i^{ts}$ \;
$\alpha_i \leftarrow \alpha_i + 1$\;
}{
$\beta_i \leftarrow \beta_i + 1$\;
}
}
\caption{Thompson Sampling}\label{alg:ts}
\end{algorithm}
 
Thompson sampling (TS) is an MAB algorithm which maintains a beta distribution 
for the reward of each arm \cite{mabts1933,NIPS2011_4321}, 
where arms are pulled randomly
according to their probabilities of being optimal.

The idea is that the reward is based on the number of 
direct choices made from a variable selected at node $j$ in the search tree.
The rationale is to make the reward more position dependent compared with
sub-tree size which varies depending on position.
However, we employ the usual 2-way branching for search
(i.e. left branch ($x = a$), right branch ($x \neq a$))
so the node degree is not the desired number of children.
We define the ({\em effective}) {\em number of children} of node $j$, $C(j)$
as follows. When node $j$ fails, $C(j)$ is the number of left branches along the right most
failed path in the sub-tree from node $j$.

To make larger rewards better,
we take the inverse value of $C(j)$ to be the reward $R^{ts}(j)$ of the heuristic at node $j$:
\begin{equation}\label{equa:tsreward}  
R^{ts}(j) = 1/C(j) 
\end{equation}

Algorithm~\ref{alg:ts} is the TS algorithm applied
to our MAB adaptive search framework,
with a simple an efficient implementation.
The functions \initTS(), \selectTS(), \updateTS() correspond to the 
three steps in the framework.

In \initTS(), we initialize the two parameters $\alpha_i$ and $\beta_i$ for each MAB selector to be 1, so the beta distribution starts as a 
uniform distribution. 
The mean rewards $R_i^{ts}$ and best rewards $R_{best}^{ts}[i]$ of each arm 
$i$ are initialized to $0$.

We call \selectTS() to select the heuristic before exploring a search node.
In \selectTS(), we draw a sample from each arm's beta distribution and choose the arm with maximum sample value.
Once the arm is selected, the algorithm applies the selected heuristic to instantiate variables and explore the search node.

When a backtrack happens, we compute the reward $r$ for arm $i$, and update the beta distribution of arm $i$ in function \updateTS(i, r).
We compare an arm's current reward with its best reward seen so far (line~\ref{alg:ts:l1}).
If $r$ improves or equals the current best reward $R_{best}^{ts}[i]$ of arm $i$, i.e.~arm $i$ explored the fewest number of children at current node,  
we consider as a success trial and increase $\alpha_i$ in the $Beta$ 
distribution by $1$, 
otherwise it is a failed trial, increasing $\beta_i$ by $1$. 

\subsection{UCB1 adaptive search}\label{sec:ucb1} 
 
UCB1 \cite{mabucb12002} is designed to give 
an expected logarithmic growth of regret. 
In UCB1, the MAB selector pulls the arm, arm $i$, 
which maximizes the value of $\rho(i)$ according to the following function:
\begin{equation}\label{ucb1}
\rho(i) = R_i + \sqrt{2 log(m)/m_i}
\end{equation}

In Equation \ref{ucb1}, $R_i$ is the mean of the past reward of arm $i$, $m_i$ is the number of past trials of arm $i$ and $m$ is the total number of trials that have been done.
So the first term $R_i$ promotes the arm gaining more rewards in the past, 
while the second term is for exploration by encouraging the arms which 
have been less frequently applied.

Typically, each constraint in a CSP can be thought of as mapping to
a propagator in the solver and each propagator has a certain level of
consistency, e.g., generalised arc consistency, bounds consistency, etc.
Since the size of search tree, the number of explored nodes, can
dominate the solving time of solving a CSP solutions where the consistency level of propagation is fixed, we define a reward function which depends 
on the ability of the heuristic (arm) in reducing the search space.
The reward $R_i(j)$ for arm $i$ at search node $j$ is defined as:
\begin{equation}\label{equa:mabreward}  
R_i(j) = 1 - C(j)/{max_{m=1..j}(C(m))} 
\end{equation}
Our reward is inspired by ~\cite{paparrizou2015} but 
uses the number of effective children of a node versus 
CPU time of sub-tree and a uniform selector.

In our framework for UCB1, the mean reward $R_i$ is initialized as 0  in 
the \initTS() procedure.
Before backtracking, all candidate heuristics are selected in a 
round robin fashion, because rewards are only updated when a backtrack happens.
This setting follows the second term of UCB Equation (\ref{ucb1}).
There is also the possibility of customizing the initial mean rewards 
of different arms to make the selection biased towards some heuristic 
in cases where certain heuristics may be known to give good results for 
certain problems.
Before exploring a search node, the MAB arm-select procedure  \selectTS() 
selects an arm which maximizes $\rho(i)$ in Equation (\ref{ucb1}), 
then the chosen heuristic from the arm is used to explore the tree node. 
When search backtracks from node $j$, the rewards of the heuristic $i$ 
used at $j$ is computed using Equation (\ref{equa:mabreward}) 
as in procedure \updateTS() to update the mean reward of arm $i$.

\subsection{Dynamic UCB1 and TS search}\label{sec:ts}
The TS and UCB1 algorithms are meant for when the distribution of rewards 
during search are fixed, i.e. a stationary probability distribution. 
We can take the view that rewards could vary over time during search, thus,
we propose to apply a non-stationary form of TS and UCB1, which consider the  rewards of the most recent $K$ search nodes dynamically. 
A non-stationary form of the UCB algorithm, sliding-window UCB, 
was proposed in ~\cite{swucb2011}. We also apply the sliding-window 
form of UCB to TS.
This gives us two dynamic adaptive heuristic variants
with window size $K$: UCB1-$K$ and TS-$K$.

In our UCB1-$K$ (TS-$K$) algorithm during search, we first check the number of explored search nodes. When this number is less and equal than $K$, UCB1-$K$ (TS-$K$) is the same as UCB1 (TS); and when the number is greater than $K$, we only take  the most recent $K$ search nodes into consideration. Specifically, for UCB-$K$, the value of $R_i$ and $m_i$ in Equation (\ref{ucb1}) are based on the recent $K$ nodes, and $m$ equals $K$. 
Similarly, for TS-$K$, the value of $R_{best}^{ts}[i]$, $\alpha_i$, and $\beta_i$ are also from the recent $K$ nodes. Note that $R_i$ and $m_i$ of UCB1-$K$ can be updated in constant time, while 
$R_{best}^{ts}[i]$ needs a priority queue to find the minimum within logarithmic time in the worst-case. Thus the overhead of TS-$K$ is more that of the UCB1-$K$. 
 
\section{Experiments} 
\label{sec:exp}

We present experiments to evaluate our MAB inspired search
heuristics.  
Our aim is to investigate if the MAB-based heuristics are more robust than
the original ones. We also want to investigate the overall performance
of all the tested heuristics.
We compare our MAB-based methods with
the candidate variable search heuristics used
as choices with the MAB algorithms.
Thus, the candidate search heuristics are one baseline.
We also compare with another straightforward baseline stochastic strategy, 
\textit{random-arm}, which chooses one heuristic from the candidate ones  
randomly at each search node.
Note that random-arm is different from a pure random heuristic which
instantiates variables randomly.  
In the TS and UCB1-based methods, we employ only a single MAB selector 
for the whole search tree.
An alternative is to have multiple MAB selectors for each search level. 
Preliminary experiments on MAB with multiple search level selectors
did not show them to be superior to a single selector.
We have omitted the results due to lack of space.
In the evaluation, the TS and random-arm methods are stochastic, while
UCB1 and the four baseline heuristics are deterministic.

\begin{table*}[!th]
\centering
\small
\setlength{\tabcolsep}{1.5pt}
{
\begin{tabular}{|c|c|r|r|r|r|r|r|r|r|r|r|r|} \hline
   \multicolumn{2}{|c|}{  }   &  \scriptsize{ddeg/dom} & \scriptsize{wdeg/dom} & \scriptsize{impact} & \scriptsize{activity}& \scriptsize{UCB1}  &  \scriptsize{UCB1-100} &\scriptsize{UCB1-500} &\scriptsize{TS} &  \scriptsize{TS-100} &\scriptsize{TS-500} & \scriptsize{random-arm} \\\hline
\multicolumn{2}{|c|}{\scriptsize{\#solved instance}}  & 311 &311&314& 311&323& 318 &316&324&322&\textbf{328}& 317 \\\hline
 {\scriptsize{runtime}}  &{ \scriptsize{mean}}  & 44.3&43.3&13.9&6.4&\textbf{3.5}&7.9&7.2&15.4&8.3&5.1&12.6\\ 
{\scriptsize{ratios}} &  \scriptsize{standard dev.} & 332.1&331.5&31.3&25.4&\textbf{9.1}&40.2&34.9&96.6&34.9&12.2&53.5 \\ 
{\scriptsize{to}} &  \scriptsize{geomean} & 2.3&2.5&4.8&2.3&\textbf{2.1}&2.4&2.3&3.1&2.9&2.7&3.5 \\
 {\scriptsize{VBS}} &  \scriptsize{maximum} & 4607.8&4619.1&298.7&308.4&\textbf{120.7}&553.9&452.7&1348.6&416.5&136.6&516.9 
\\\hline
\end{tabular}
} 
\caption{Overall results for all search heuristics.\label{exp-overall}}
\end{table*} 

We evaluate our search heuristics on a variety of
structured and unstructured problems, 
to investigate the search behavior across a range of problems.
The benchmarks are from the CSP solver competition 
({\small \tt \url{http://www.cril.univ-artois.fr/CSC09}}). 
We use 363 problem instances from 15 problem series.\footnote{Instances that are solved with no search or those where all heuristics timeout are ignored.
Note that applying SAC at the root node can solve some problems
without search.
} 
The structured problems are: traveling salesman (TSP), costas array, resource constrained project scheduling (RCPSP), balanced incomplete block designs (BIBD), all-interval, golomb ruler, crossword, FPGA, ssa and modified-renault.  
The unstructured problems are the hard random ones: rand3-20-20, rand3-20-20-fcd, rand8-20-5, dagrand, and half. 
The benchmark CSPs are all non-binary (but can have some binary constraints) 
and chosen to have diverse constraints, 
including extensional (table), intensional and also global constraints.
The experiments were run on a 3.40GHz Intel i7-4770 machine.

\begin{figure}[!th]
\centering
\hspace*{-5cm}
\includegraphics[width=0.42\textwidth]{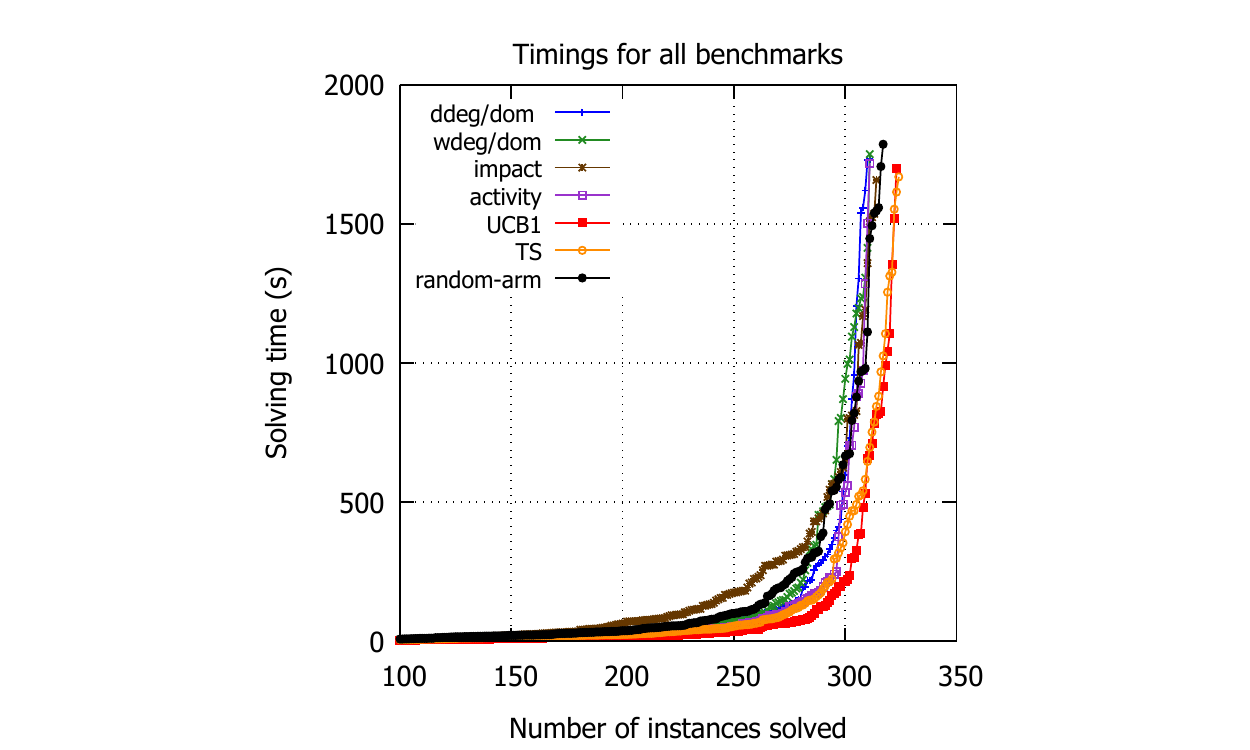} 
\hspace*{-3.3cm}
\includegraphics[width=0.42\textwidth]{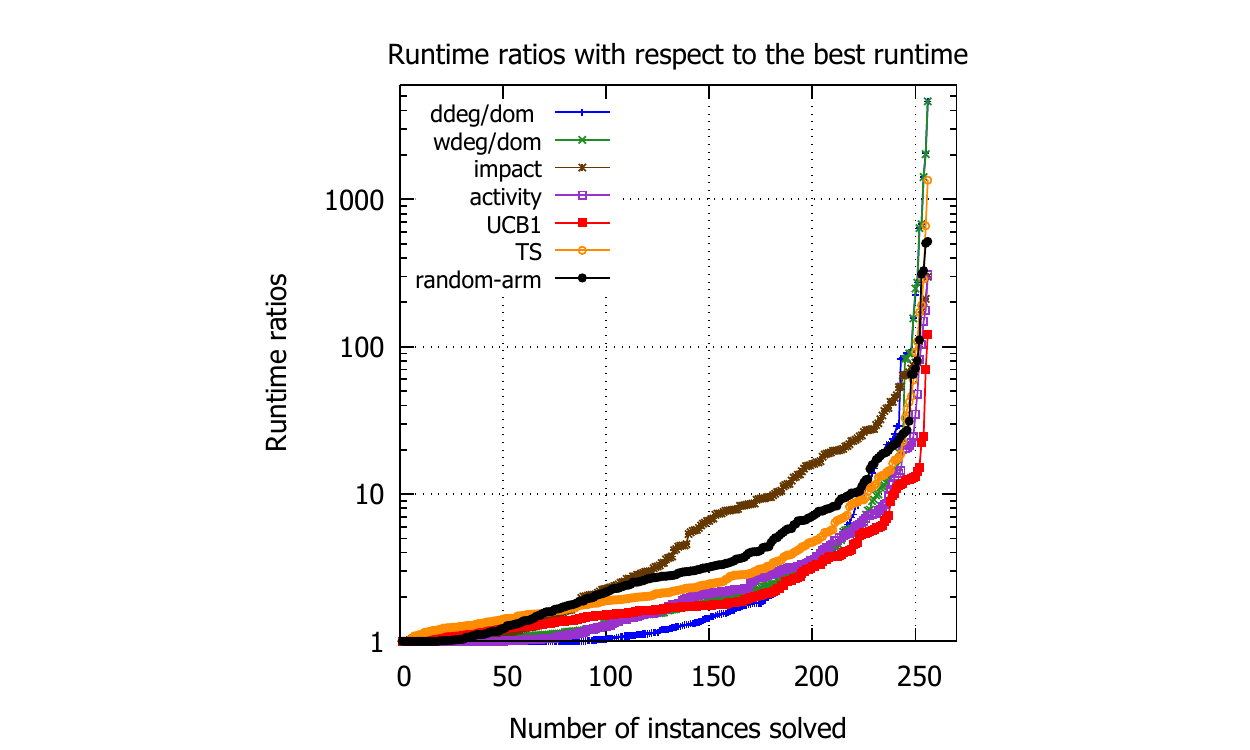} 
\hspace*{-5cm}

\hspace*{-5cm}
\includegraphics[width=0.42\textwidth]{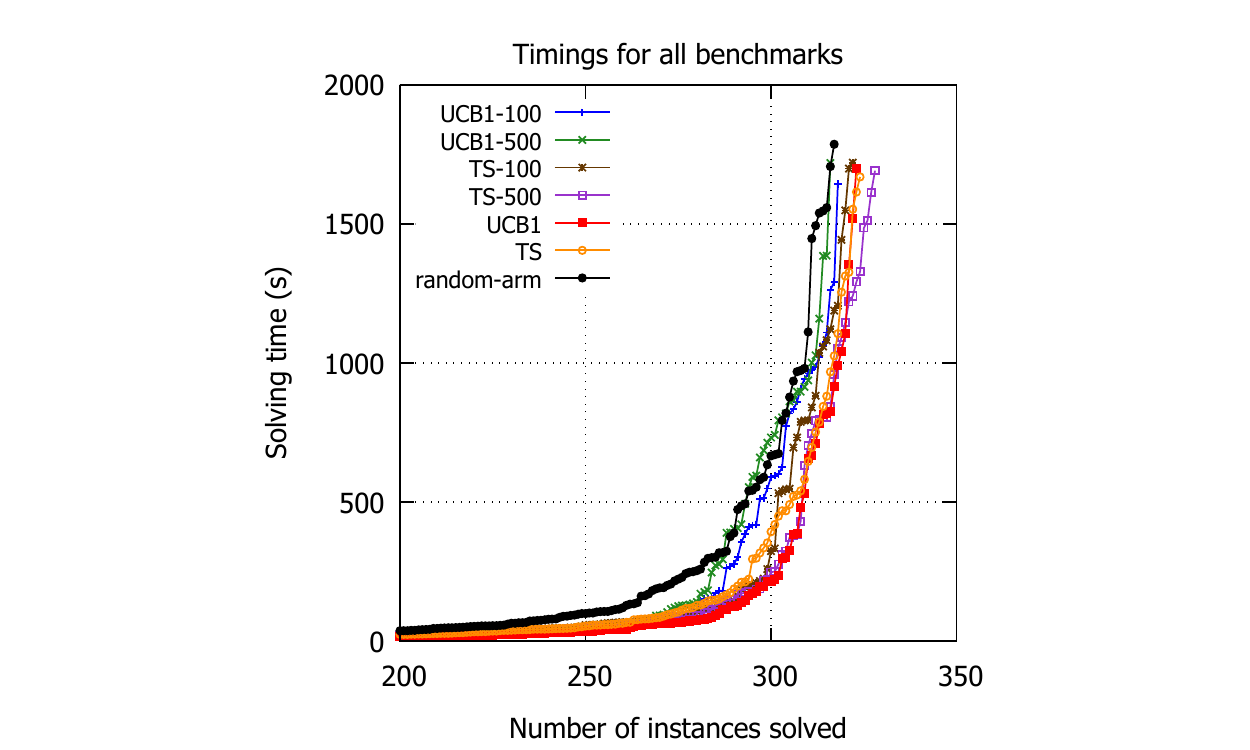} 
\hspace*{-3.3cm}
\includegraphics[width=0.42\textwidth]{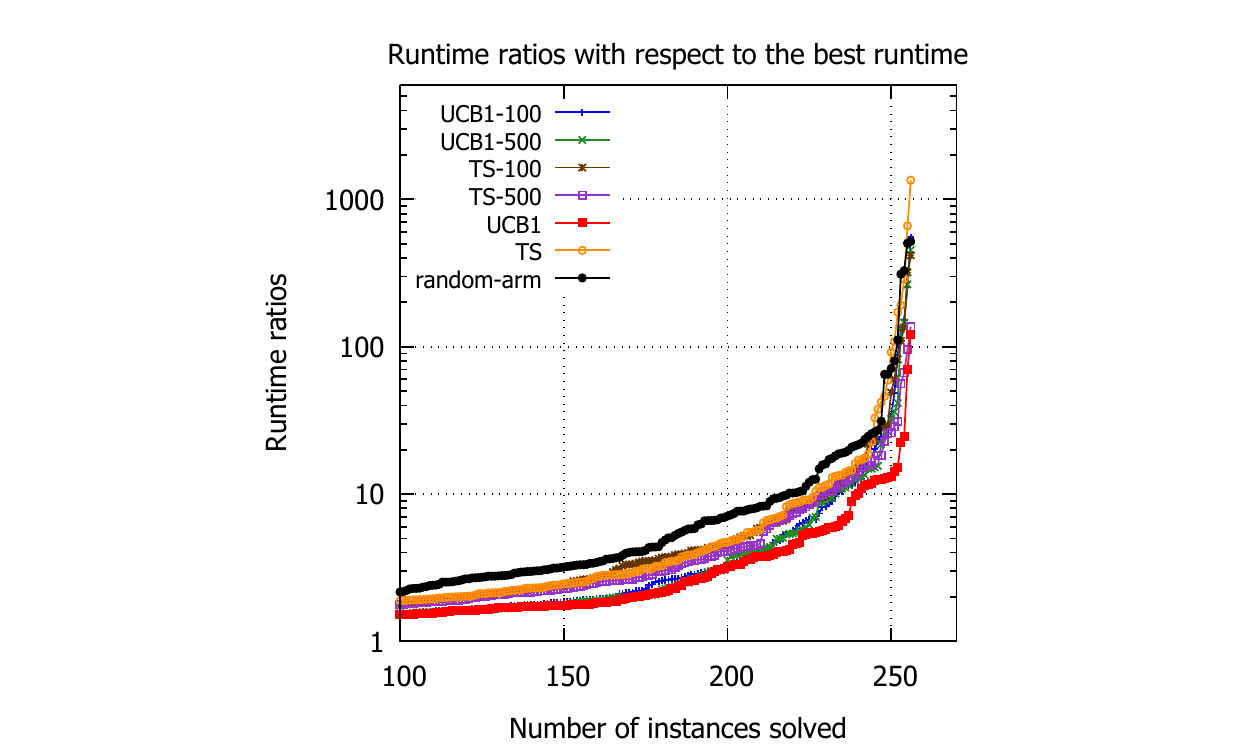} 
\hspace*{-5cm}
\caption{Distribution of runtime and runtime ratios to VBS.  \label{fig:graph1}}
\end{figure}

We use the AbsCon solver ({\small \tt \url{https://www.cril.univ-artois.fr/~lecoutre/software.html}}) 
for its versatility as a blackbox solver: 
many propagation algorithms and heuristics are implemented and selectable. 
We focus on search heuristics and their relative effectiveness in the experiments, thus the consistency levels and propagators for constraints are 
the AbsCon defaults.\footnote{The default consistency is Generalized Arc Consistency.} 
We employ a full initialization of variable impact and activity at the root node of the search tree using the singleton arc consistency~\cite{sac2000} propagator SAC3~\cite{sac32005} in AbsCon which is needed to
initialise the activity and impact heuristics. 
In order to ensure that search starts from the same state for different solving strategies, we apply the SAC3 propagation at the root node for all methods. The overhead of this initialization is negligible. 
CPU time is limited to 1800 seconds per instance and
memory to 8GB.

In our experiments, we employ the four well-known and commonly used variable ordering heuristics (discussed previously): $ddeg/dom$, 
$wdeg/dom$, \textit{impact}, and \textit{activity}. 
We use these heuristics as the candidates (arms) of the MAB methods
and also for random-arm.
As we focus on investigating variable heuristics, we use the 
same lexicographic value heuristic (\textit{lexico}) for
all cases.

\subsection{Overall results}
 
To investigate robustness, we can measure performance with respect
to the best runtime (Virtual Best Solver (VBS)) per instance as
the runtime ratio to VBS, i.e. runtime/(VBS runtime).
In order to compute the runtime ratios of all heuristics, we ignore 
an instance if there is one evaluated heuristic which cannot solve the instance 
within the timeout.
Thus, the runtime ratios are computed based on 256 
(out of 363) instances that are solved by all heuristics.

Table~\ref{exp-overall} gives the overall statistics (arithmetic and geometric
mean, standard deviation, maximum) of all search heuristics
using their runtime ratios.
We see that UCB1 is the most robust with the smallest
standard deviation and maximum ratio with respect to the VBS runtime. 
UCB1 also has the smallest mean ratio of 3.5 to VBS.
The maximum ratio of UCB1 is 120.7, which is about 38X smaller than \textit{ddeg/dom} and \textit{wdeg/dom}. 
This shows that the baseline heuristics can give highly variable results
highlighting the importance of robust heuristics.
We see that TS-500 solves the most problems, slightly more than UCB1,
but has higher means and standard deviation.

The graphs in Fig.~\ref{fig:graph1} show the 
overall runtime distribution. The top two graphs are for individual heuristics and non-dynamic MAB-search while the bottom two graphs are 
for the dynamic MAB-search methods.
The graphs on the the left are based on all 363 instances as they use solving
time while the graphs on the right use the runtime ratio to VBS
and are based on 256 instances solved by all search heuristics.
Each point (x, y) in the left graphs shows that the technique solves x 
instances within y seconds\footnote{
	Note that the y coordinate represents the runtime of each individual x instance, not the total runtime of all x instances.
}
while each point (x, y) in the right graphs shows that the technique solves x instances within y times of the VBS runtime. 
From Table \ref{exp-overall}, we saw that UCB1 had good robustness,
the runtime distribution in Fig.~\ref{fig:graph1} show that
UCB1 (red line) also has the best overall result for the majority of
instances.
We highlight that the MAB methods have higher overheads
as they also include the overhead of maintaining the rewards of the 
heuristic taken at each search node, as well as the variable scores of the unselected heuristics, whereas the underlying heuristics do not have this overhead.

The runtime of the TS method is also robust.
In both of the graphs in Fig.~\ref{fig:graph1}, the lines for TS 
are closer to the best search strategies compared with the worst ones. 
We also see a surprising result. 
The simple random-arm heuristic is not the worst strategy, 
which might not be expected a priori, and can beat some of the original
baseline heuristics.
We observe that the random-arm method choosing among the baseline heuristics 
results in some robustness but as it does not have any exploitation,
it has worse overall performance.

\begin{figure*}[tb]
\centering
\hspace*{-5cm} 
\includegraphics[width=0.45\textwidth]{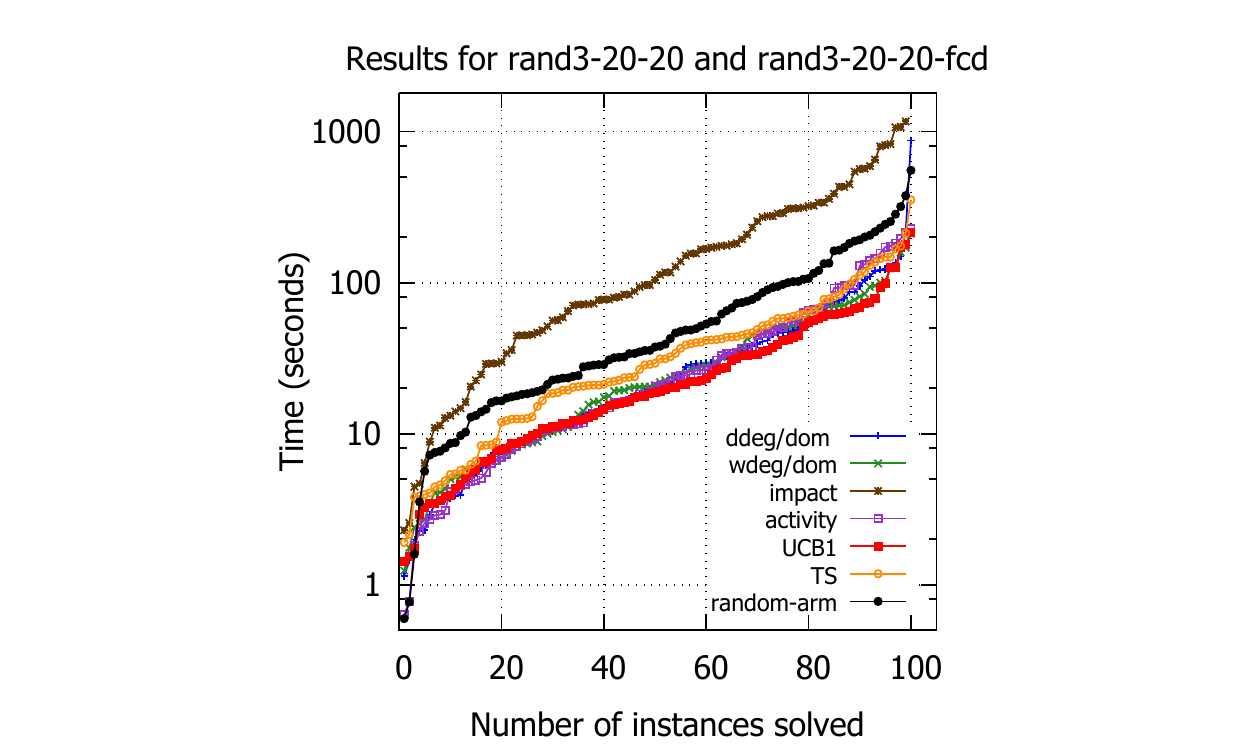}\hspace*{-3.7cm}
\includegraphics[width=0.45\textwidth]{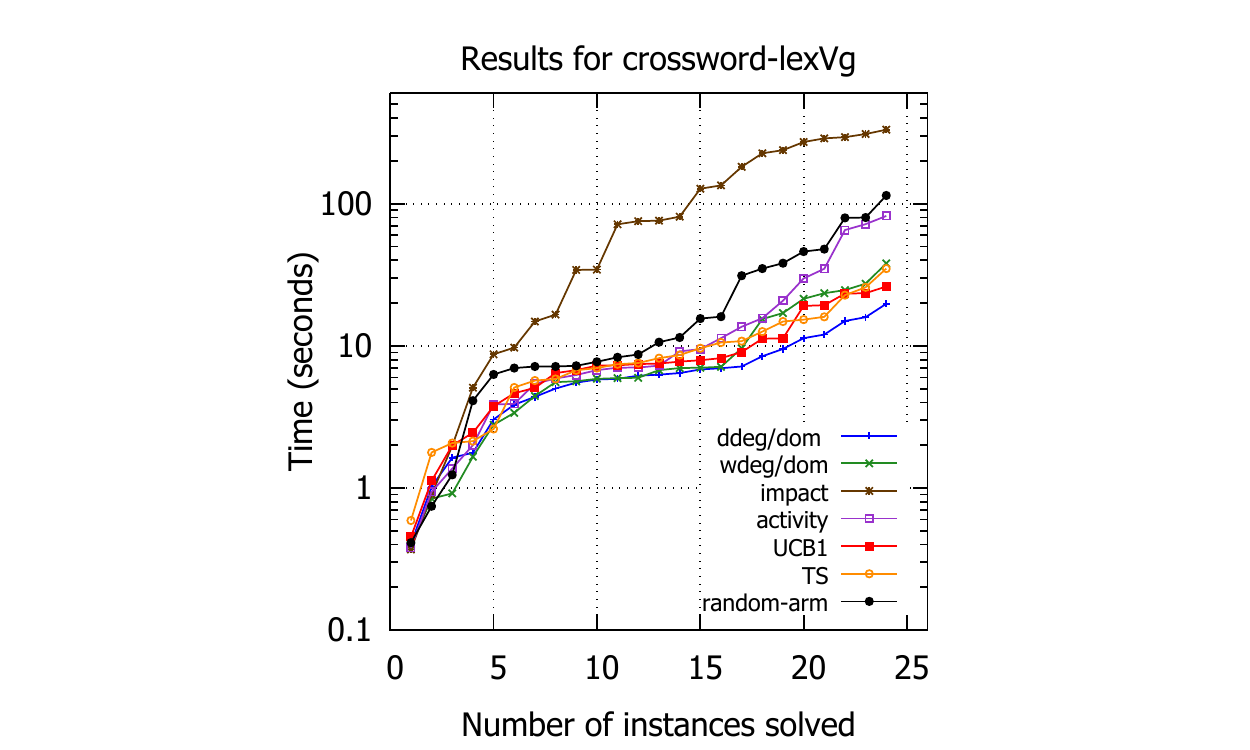}
\hspace*{-3.7cm}          
\includegraphics[width=0.45\textwidth]{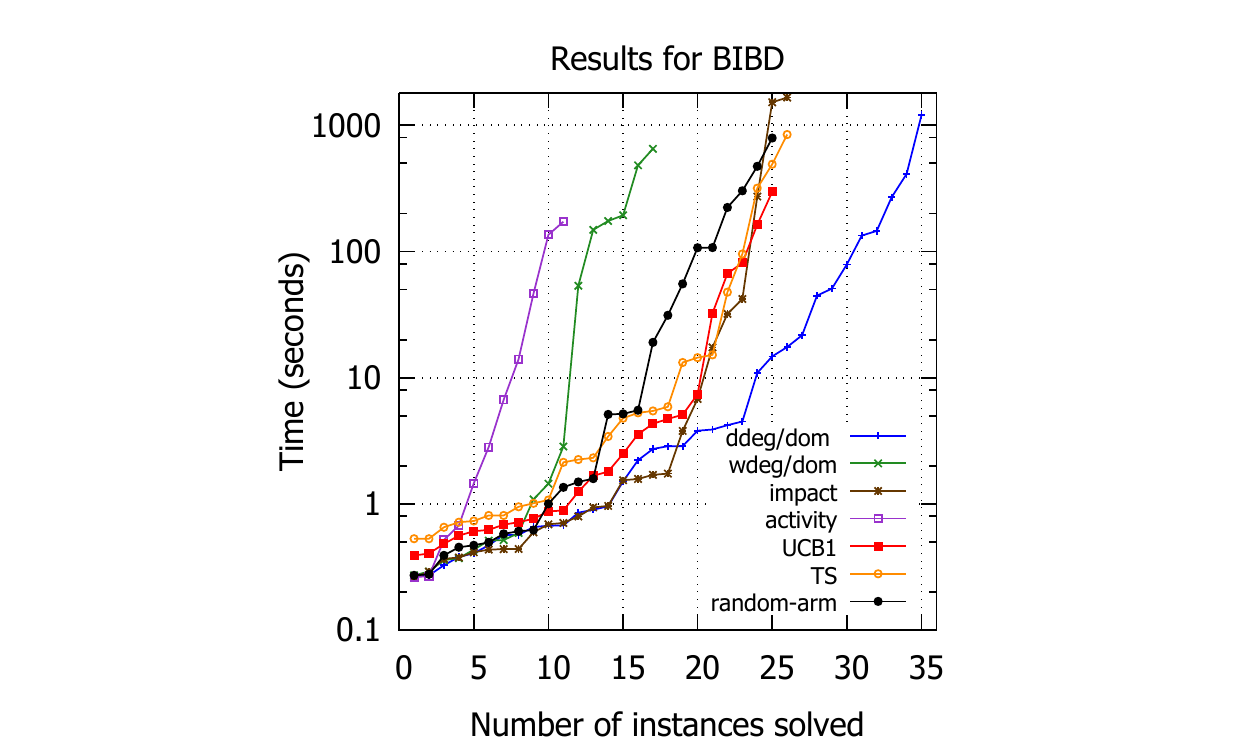} \label{fig:subb1}
\hspace*{-3.7cm}    
\includegraphics[width=0.45\textwidth]{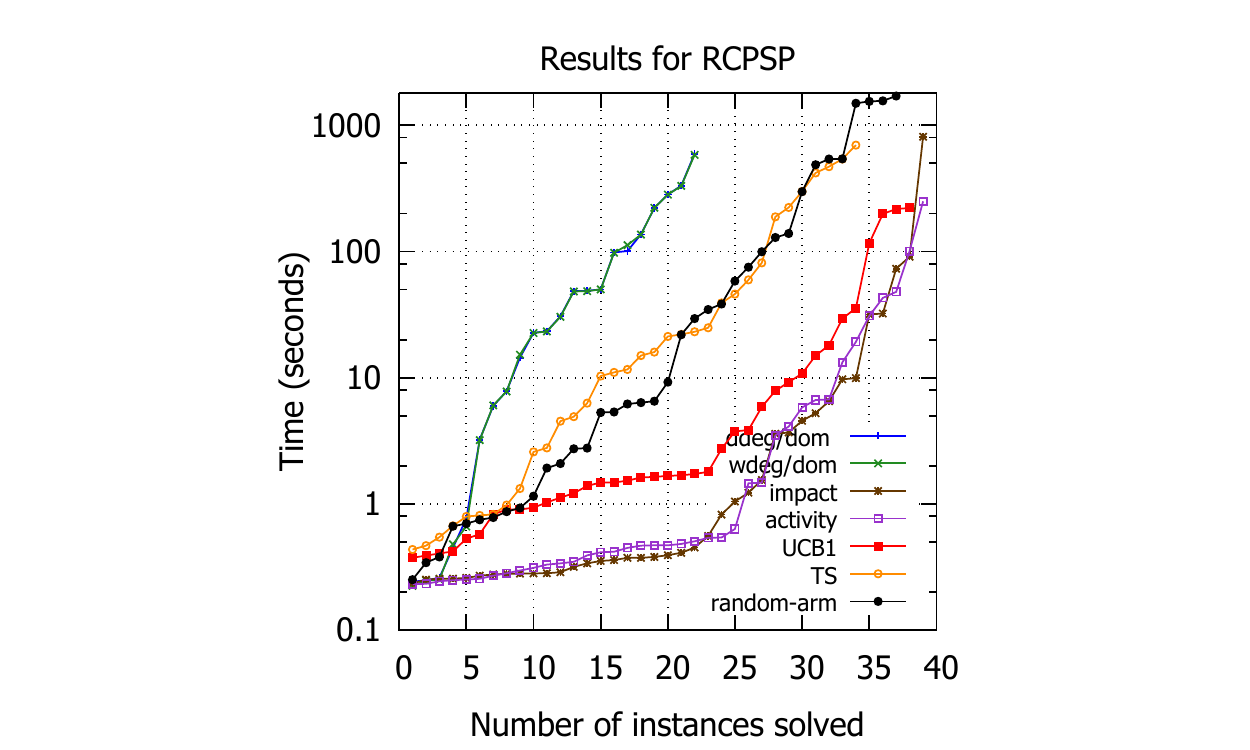}       
\label{fig:subb2}  
\hspace*{-5cm}
\caption{Runtime distribution for benchmark series rand3-20-20-fcd, dagrand, crossword-lexVg, RCPSP, BIBD and FPGA. All FPGA instances timeout under the $ddeg/dom$ heuristic, thus there are no points in the FPGA graph for $ddeg/dom$.\label{fig:graph2}}
\end{figure*}

\begin{figure*}[tb] 
\centering
\begin{subfigure}[t]{1.0\textwidth}{
\hspace*{-5cm}
	\includegraphics[width=0.42\textwidth]{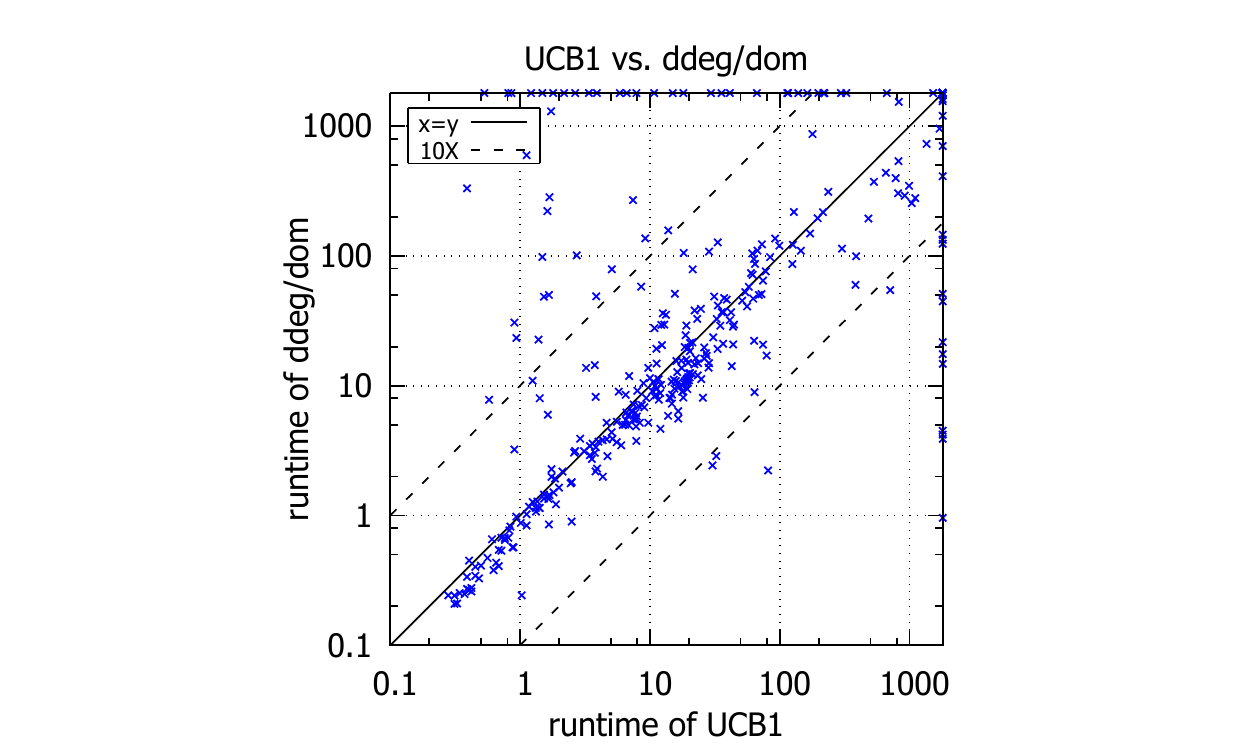}\hspace*{-3.5cm} 
	\includegraphics[width=0.42\textwidth]{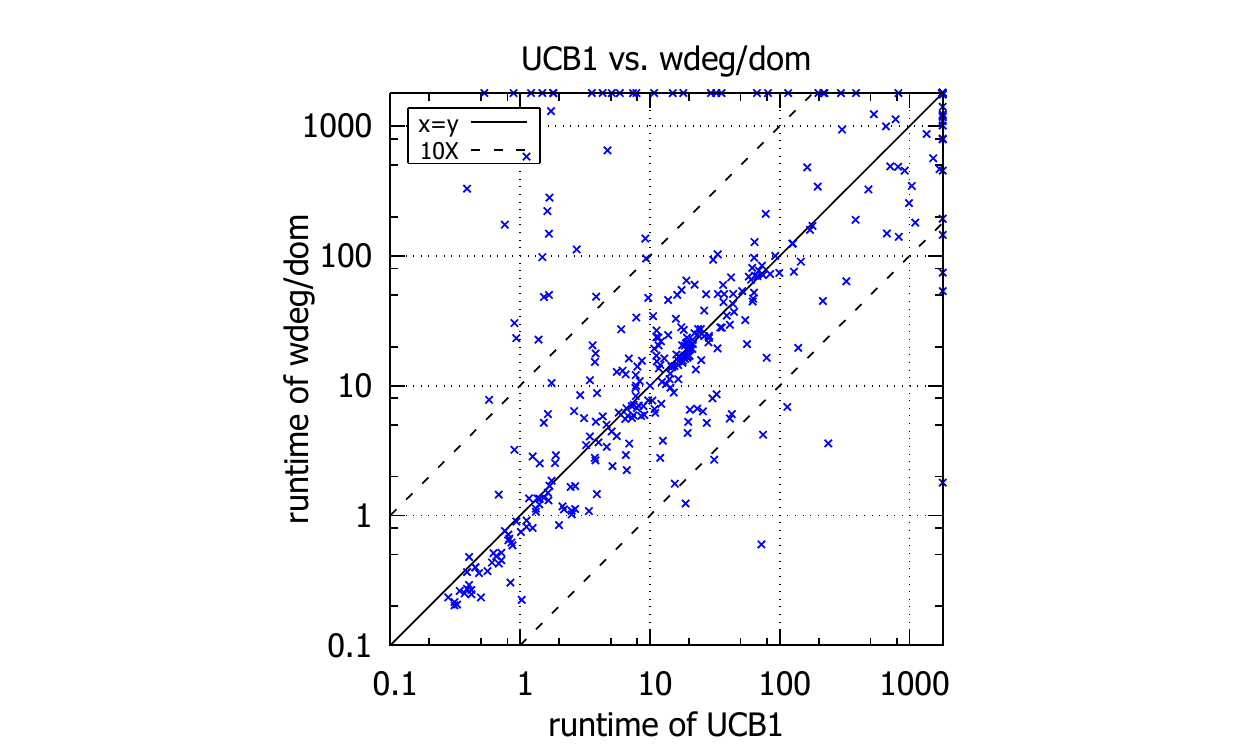}
\hspace*{-3.5cm}
	\includegraphics[width=0.42\textwidth]{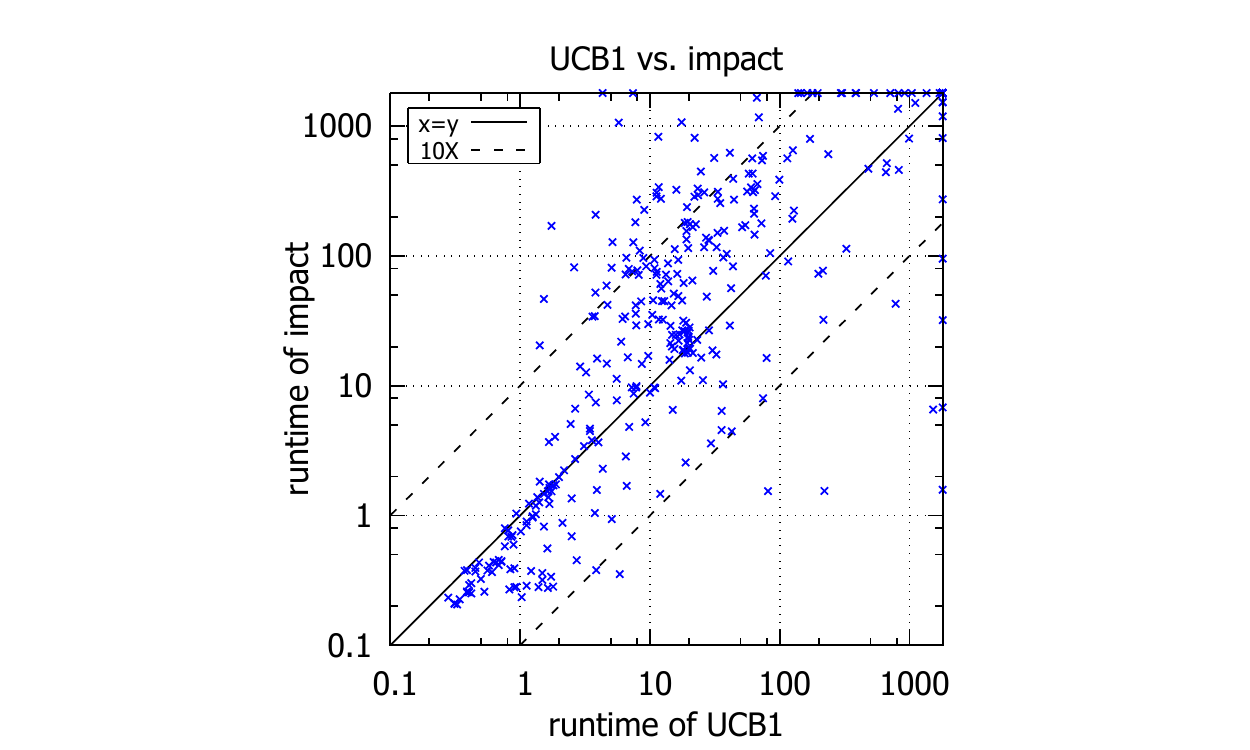}\hspace*{-3.5cm} 
	\includegraphics[width=0.42\textwidth]{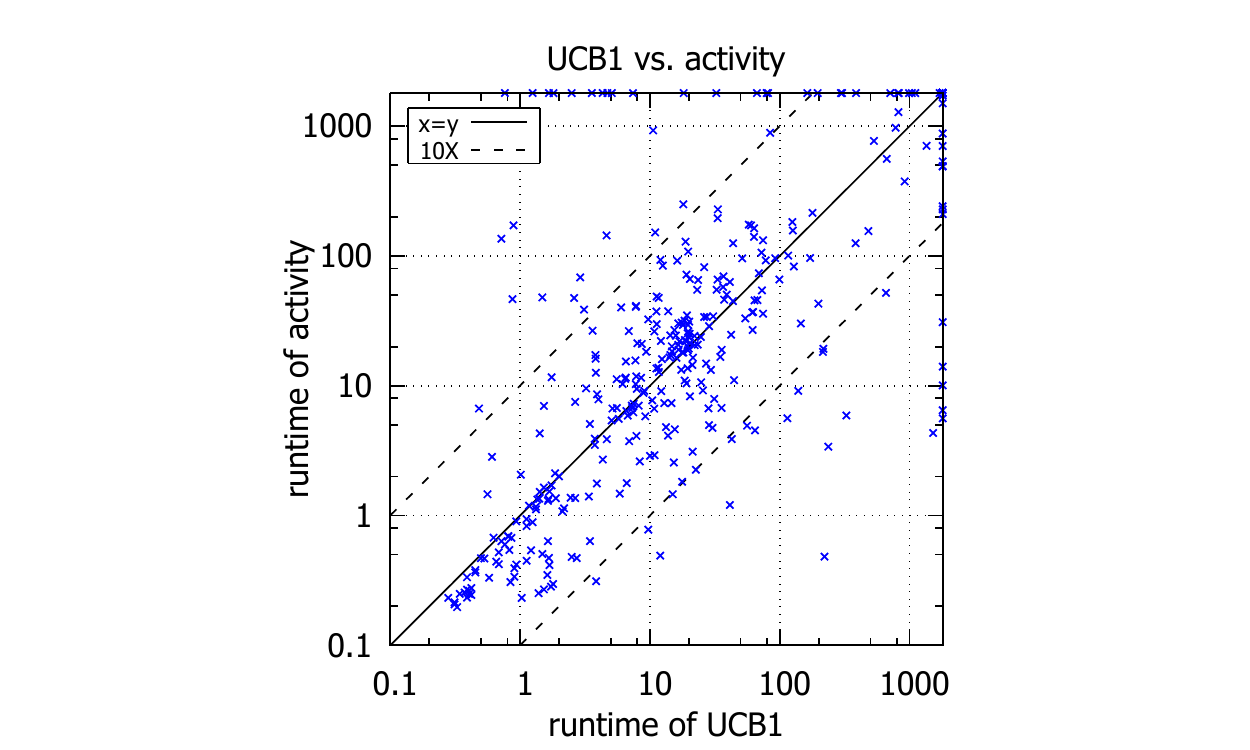}
	\hspace*{-5cm} }\centering \caption{UCB1 vs. individual candidate heuristics}\label{fig:graph3:a}
	\end{subfigure} 
\begin{subfigure}[t]{1.0\textwidth}{	
	\hspace*{-5cm}
	\includegraphics[width=0.42\textwidth]{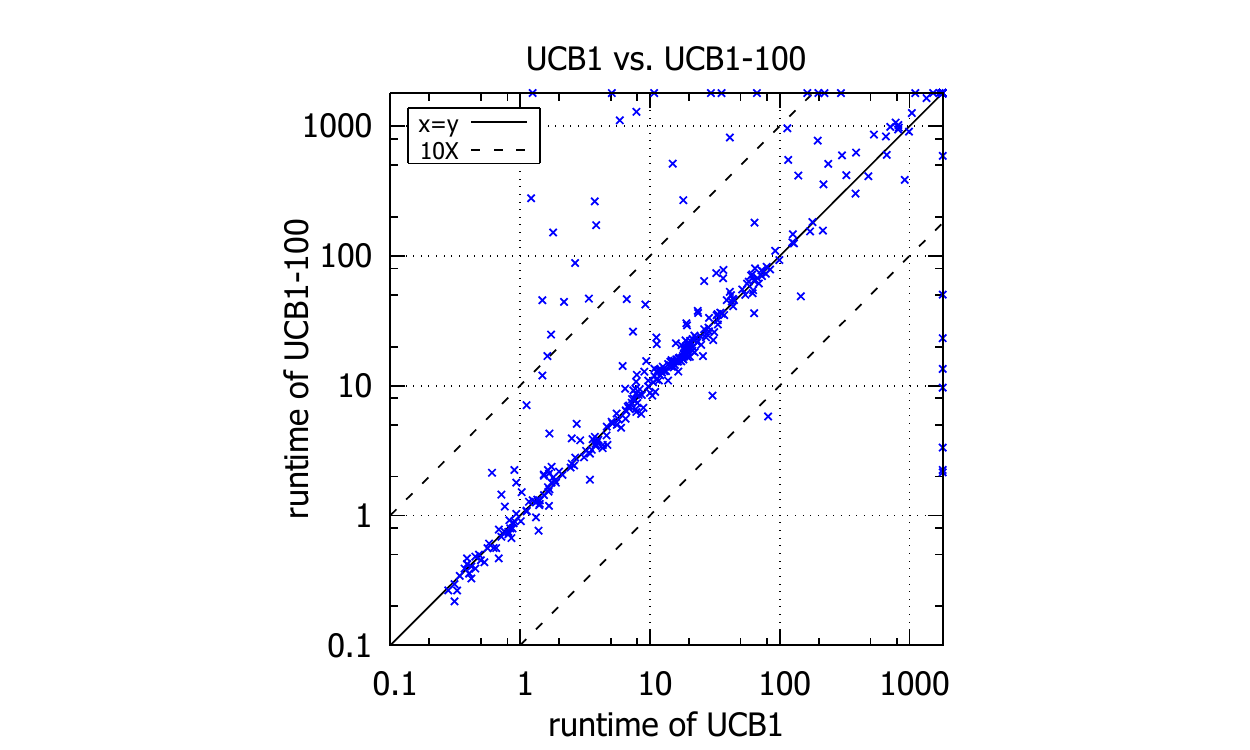}\hspace*{-3.5cm} 
	\includegraphics[width=0.42\textwidth]{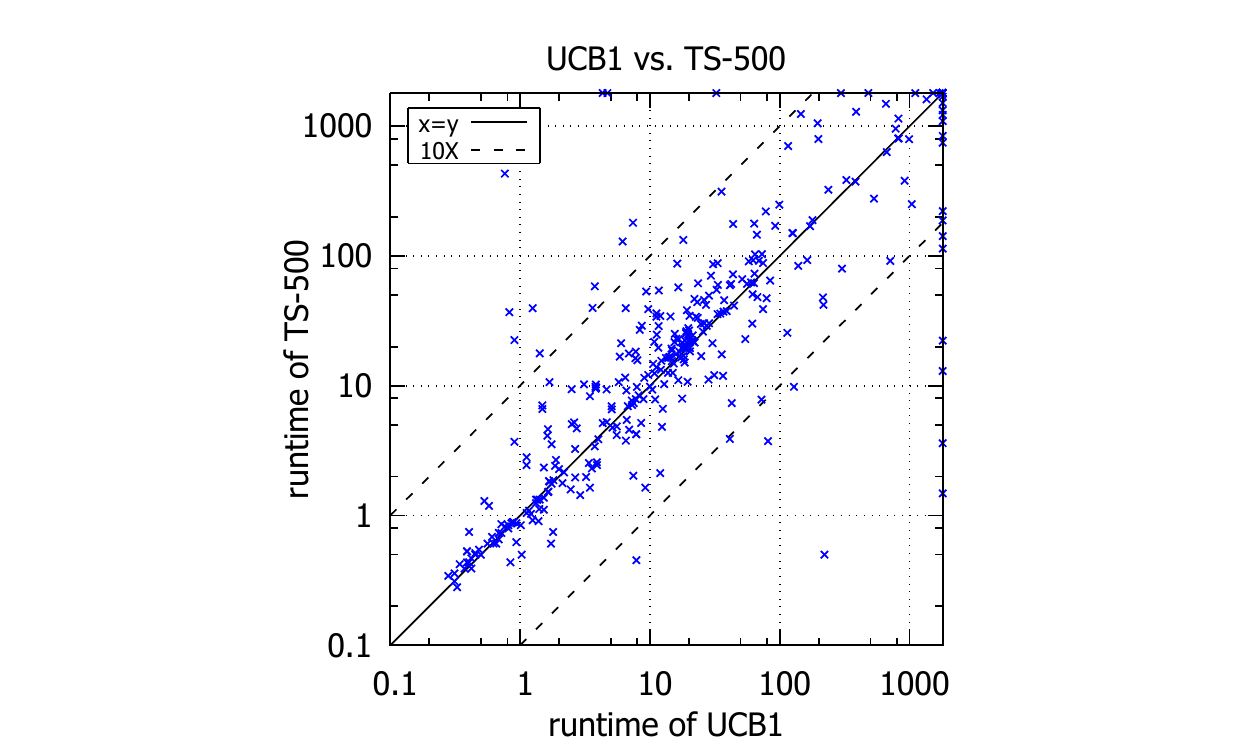}
\hspace*{-3.5cm}
	\includegraphics[width=0.42\textwidth]{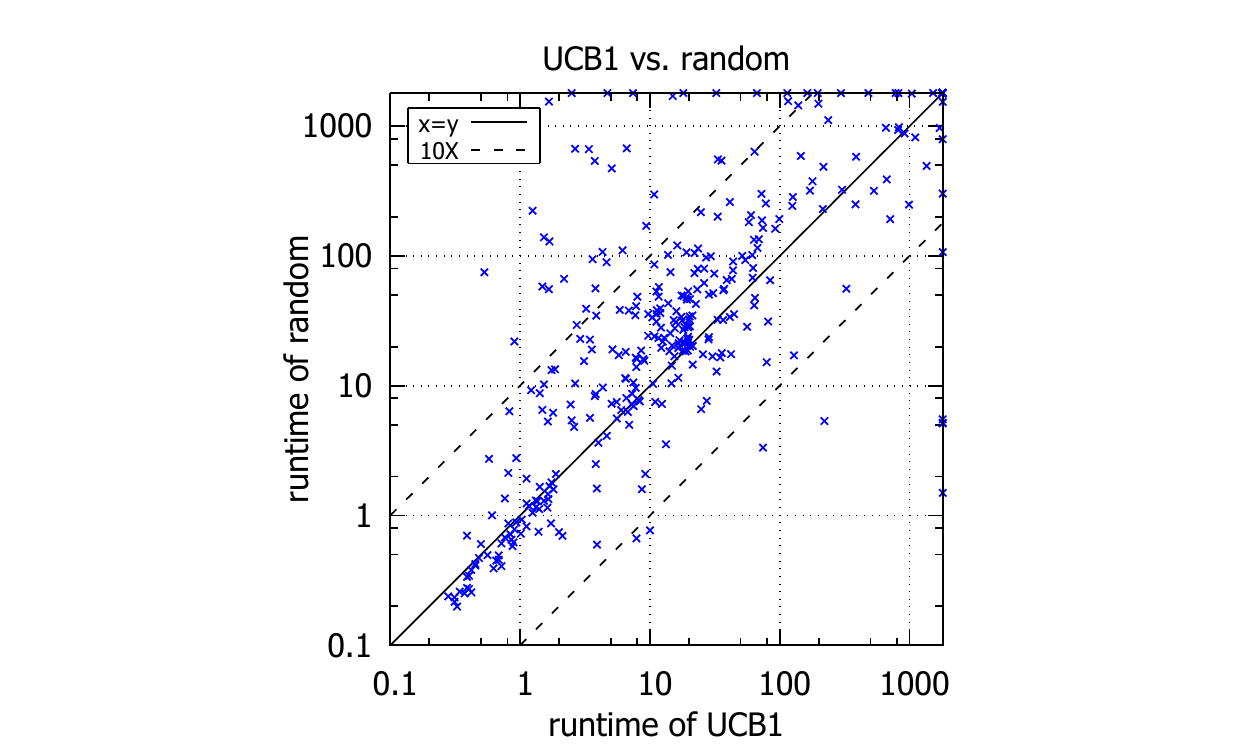}\hspace*{-3.5cm} 
	\includegraphics[width=0.42\textwidth]{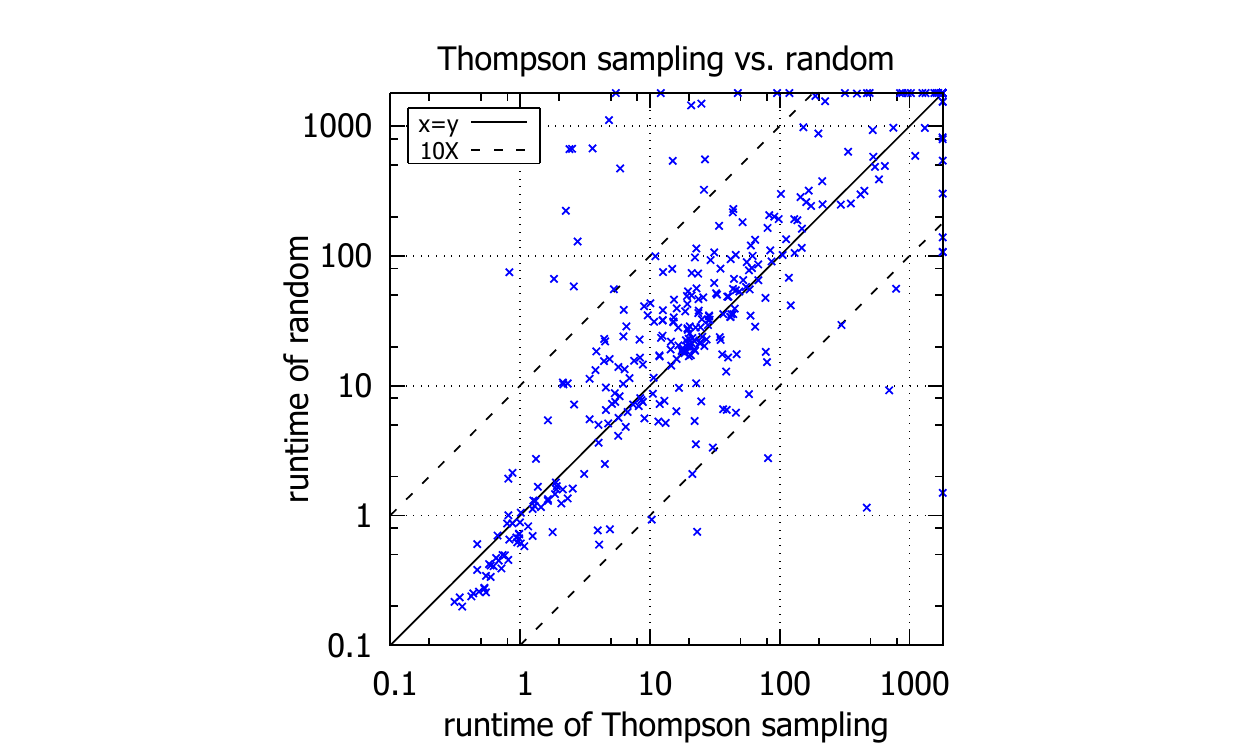}
	\hspace*{-5cm} 	
	\label{fig:graph3a}}\centering \caption{Comparison of MAB search strategies} \label{fig:graph3:b}	
\end{subfigure}     
\caption{Runtime comparison of MAB methods, individual candidate heuristics, and random heuristic on all instances.\label{fig:graph3}}
\end{figure*}   

\subsection{Robustness by benchmark series}

The graphs in Fig.~\ref{fig:graph2} present in detail the runtime distribution 
of four specific problem series showing that the MAB methods are 
robust--especially the UCB1 strategy. 
Dynamic MAB methods are not given to avoid cluttering the
graph as they were not as good overall as UCB1 on these series. 
Fig.~\ref{fig:graph2} illustrates that the baseline heuristics while being
good on some series are not robust.
For example, if we ignore the UCB1 method, $ddeg/dom$ and the $wdeg/dom$ are the best two variable heuristics for rand3-20-20 and rand3-20-20-fcd, and the worst is the $impact$  heuristic.  
However for RCPSP, the result is just the opposite that $impact$ is much better than $ddeg/dom$ and $wdeg/dom$. 
We also see large differences between different heuristics for other problems.
For example, in BIBD, the $ddeg/dom$ heuristic is faster than $activity$ by several orders of magnitude, but for the FPGA problem, $ddeg/dom$ does not solve any instance (graph for FPGA is not given for space reasons).
This highlights the importance of having a robust heuristic.

We compare the runtime of different methods in pairs in Fig.~\ref{fig:graph3}.
Fig.~\ref{fig:graph3}(\subref{fig:graph3:a}) compares the runtime of UCB1 
with individual heuristics.  
The points located on the top and right boundaries are 
instances which timeout 
on the individual heuristics and UCB1 respectively.
We can see that there are more points above the $x=y$ line 
including timeout points,
indicating that UCB1 is faster than the compared heuristic.
Furthermore, we see that the points in the upper portion are further away from the $x=y$ line than the points in the bottom portion, e.g.
most UCB1 points are within the 10x (dotted) line while  
$wdeg/dom$ have many points outside the 10x line.
This shows that when $wdeg/dom$ is slower than UCB1, it is much slower;
but when UCB1 is slower, the slowdown is lesser. 
We see a similar trend in the other graphs. 

Similarly, Fig.~\ref{fig:graph3}(\subref{fig:graph3:b}) gives the runtime 
of UCB1 compared with the dynamic MAB methods UCB1-100 and TS-500, 
and also UCB1 and TS compared with the random-arm.
The graphs of "UCB1 vs. UCB1-100" and "UCB1 vs. TS-500" show that UCB1 is better than the dynamic UCB1-100, but close to TS-500.
The graphs of UCB1 and TS versus random-arm
show that learning is effective.
Note that UCB1 is deterministic while TS and random-arm are stochastic.
 
\subsection{The frequency of candidate heuristics}

\begin{figure}[tb]
\centering
\hspace{-2cm}
         {\includegraphics[width=0.5\textwidth]{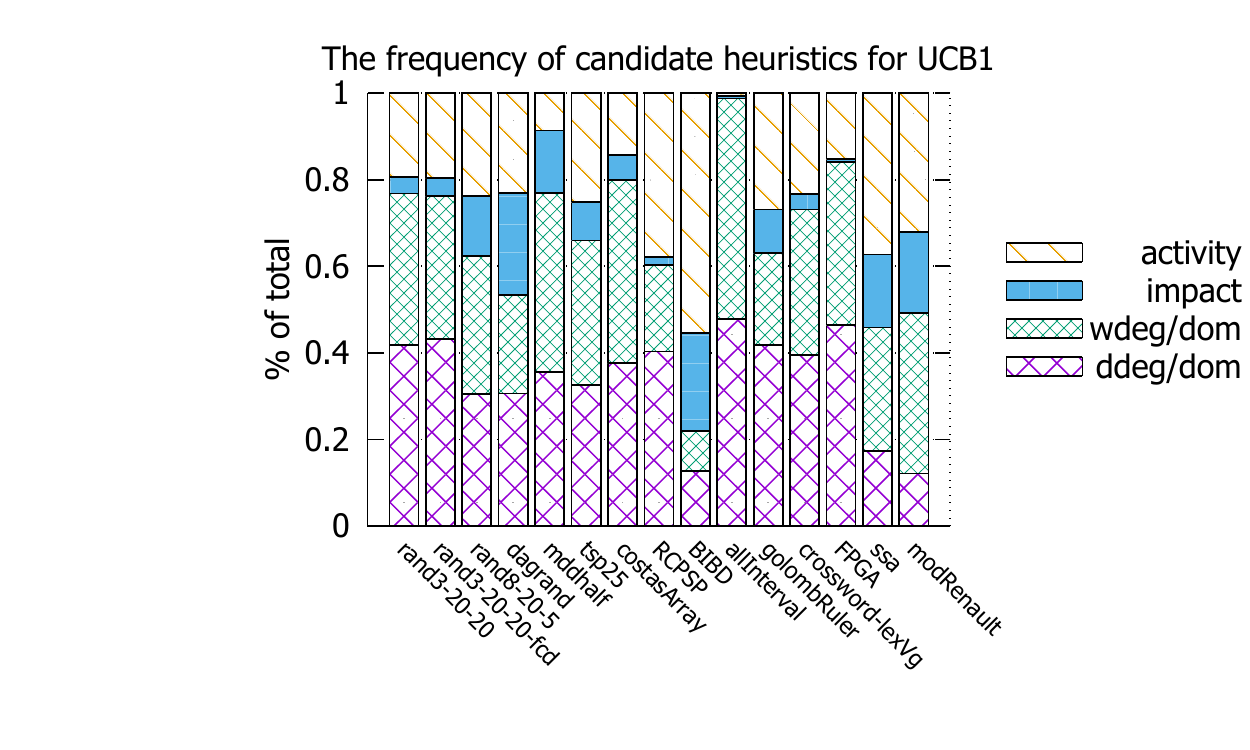}
\vspace{-2mm}
\vspace*{-3mm}
          
\hspace{-2cm}           
	\includegraphics[width=0.5\textwidth]{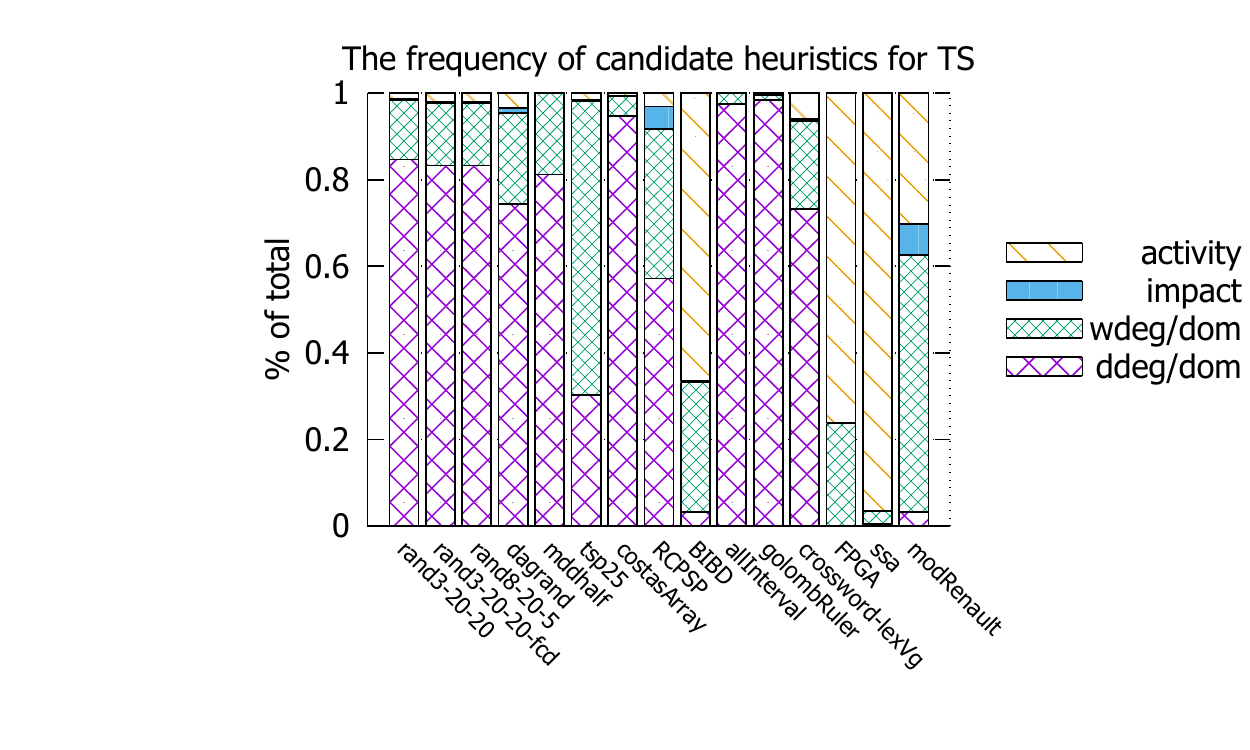}}
\label{fig:subb9}  
\caption{Frequency of candidate heuristics applied in the MAB-methods. \label{fig:graph5}}
\end{figure}

We investigate the frequency of candidate heuristics of MAB search and 
its correlation with the performance of the candidate heuristic on various
problems.
Fig.~\ref{fig:graph5} gives the mean frequency of application of each heuristic
when solving a benchmark series by UCB1 and TS.
We see that UCB1 and TS can automatically differentiate between different 
heuristics. 
A correlation can be seen between the performance of individual heuristic and 
its application frequency in the MAB-based method.
For example, in rand3-20-20 and rand3-20-20-fcd, the worst heuristic as shown in Fig.~\ref{fig:graph2}, is $impact$ which is used the least frequently.
However as our MAB-based methods are online,
such a correlation is not always the case, e.g.~for BIBD $activity$ 
is the worst heuristic but is the most frequent heuristic applied 
in UCB1 and TS.
We can also see that the frequency of heuristics used in the MAB-based
algorithms vary significantly which suggests some
complex interaction with the search process.

\subsection{Discussion on PSS}\label{sec:exp:mabpss}
Recently, parallel strategies selection (PSS) 
was shown to be a promising approach for selecting search heuristics
\cite{pss2016}.
PSS is quite different from online or supervised learning based methods.
Firstly, it exploits that a CSP can be decomposed into a large number
of sub-problems which are independent and hence parallelism can be
easily exploited.
Secondly, it uses a statistical sampling approach,
sampling sub-problems to choose the heuristic.

As PSS exploits a large parallelism factor from independent parallelism,
it is not comparable with sequential methods.
Most works on search heuristics are sequential as is this paper.
However, PSS was shown to work well, hence, we also investigate PSS although
it is not learning-based.
We implemented a form of PSS, sequential PSS (sPSS),
which is PSS with a parallelism factor of one.
This allows decomposition and sampling strategies to be compared
independently of the parallelism.

For space reasons, we summarize the results.
We found on our benchmarks that sPSS is much slower than the MAB method 
especially on unsatisfiable problems.
This is because all the sampled subproblems and the remainder subproblems
have to be solved and there is no super-linear speedup from parallelism.
Preliminary experiments show that the total number of explored search nodes 
of all subproblems of sPSS can be much more than that of MAB heuristics, e.g.~for rand3-20-20-1, the total search nodes of sPSS is 507K while the search nodes of MAB-UCB1 is 73K. This makes the sPSS much slower than MAB and also other individual search heuristics. 
We also found that sometimes the decomposed subproblems are too easy, 
with few search nodes, which makes the solver initialization overhead 
more significant in the sequential case. 
For example, the mean number of search nodes of each subproblem of the 
unsatisfiable instance ruler-34-9-a4 is 8.4, although the total number of search nodes of all subproblems is close to that of UCB1. 
As such the total runtime of sPSS is 116.8s, compared with 6.9s of UCB1. 
We also found that the performance of sPSS approach can depend greatly 
on the decomposition, which suggests that sPSS not as robust as our MAB methods.

We caution that sPSS is not PSS and comparing sequential versus parallel
algorithms is tricky.
One results illustrate the expected behavior that sequential solving
of sub-problems independently can fall into the unlucky cases that
satisfiable subproblems are only selected late in execution.
As such to benefit from the PSS approach, 
one should have sufficient parallelism, consistent with \cite{pss2016}.
We also note that our sPSS implementation is only preliminary 
and can possibly be more efficient.

\section{Conclusion}

We propose a bandit-based approach which applies various variable heuristics automatically during CSP solving. 
Unlike independent heuristics, which explores the search space only based on a 
single approach (e.g.~score function), our method considers several 
individual heuristics together and learns to apply better ones dynamically
during search in an online fashion.
Experiments show that our MAB methods are more robust than the original
heuristics and can also give better performance.

Search heuristics for CSPs has been investigated extensively, e.g.~utilizing the failures of constraints in $wdeg/dom$, or measuring the \textit{activity} of variables during propagation.
However the combination of various heuristics deserves more study as
the solving can benefit from applying different heuristics according to 
a different status of the problem during search.  
Our MAB-based learning methods shows promise in this direction.
Making a CP solver automatic and as ``black box'' as possible is highly
desirable.
Our experiments show that an automatic search strategy 
within the solver can be robust with good performance on many problems.
In contrast, the common practice for performance requires
the model or constraint program to provide a good search strategy.
However, manual search heuristic selection may require 
expert knowledge with extensive tuning effort.

It would be interesting to combine online search heuristic selection with
propagator selection. To be general, the solver should have
non-binary propagators of different consistencies which may be interesting
for global constraints.

\section{Acknowledgments}
This work was supported by MOE2015-T2-1-117.

\end{document}